\documentclass[lettersize,journal]{IEEEtran}
\usepackage{amsmath,amsfonts}
\usepackage{algorithmic}
\usepackage{algorithm}
\usepackage{array}
\usepackage{textcomp}
\usepackage{stfloats}
\usepackage{url}
\usepackage{verbatim}
\usepackage{graphicx}
\usepackage{cite}
\usepackage[colorlinks,linkcolor=red,hidelinks,]{hyperref}

\usepackage{amsthm}%Cui Zhen
\usepackage{cases}%Cui Zhen
\usepackage{bm}%Cui Zhen
\usepackage{algorithm}%Cui Zhen
\usepackage{algorithmic}%Cui Zhen
\usepackage{cases}%Cui Zhen
\usepackage{subfigure}%%cuizhen
\usepackage{multirow}%%cuizhen
\usepackage{color} %% cuizhen
\usepackage{enumerate}%% cuizhen
\usepackage{etoolbox}%% cuizhen
\usepackage{mathrsfs}%% cuizhen
\newcommand{\MyMapTemplatePrefix}[4]{\expandafter#1\csname#3#4\endcsname{#2{#4}}}
\newcommand{\MyMapTemplatePrefixNew}[5]{\expandafter#1\csname#4#5\endcsname{#2{#3{#5}}}}
\forcsvlist{\MyMapTemplatePrefix {\def} {\mathbf} {}} {A,B,C,D,E,F,G,H,I,J,K,L,M,N,O,P,Q,R,S,T,U,V,W,X,Y,Z}
\forcsvlist{\MyMapTemplatePrefix {\def} {\mathbf} {}} {a,b,c,d,e,f,g,h,i,j,k,l,m,n,o,p,q,r,s,t,u,v,w,x,y,z,1,0}
\forcsvlist{\MyMapTemplatePrefix {\def} {\widetilde} {wt}} {A,B,C,D,E,F,G,H,I,J,K,L,M,N,O,P,Q,R,S,T,U,V,W,X,Y,Z}
\forcsvlist{\MyMapTemplatePrefix {\def} {\widetilde} {wt}} {a,b,c,d,e,f,g,h,i,j,k,l,m,n,o,p,q,r,s,t,u,v,w,x,y,z} % r
\forcsvlist{\MyMapTemplatePrefixNew {\def} {\widetilde}{\mathbf} {tb}} {A,B,C,D,E,F,G,H,I,J,K,L,M,N,O,P,Q,R,S,T,U,V,W,X,Y,Z}
\forcsvlist{\MyMapTemplatePrefixNew {\def} {\widetilde}{\mathbf} {tb}} {a,b,c,d,e,f,g,h,i,j,k,l,m,n,o,p,q,r,s,t,u,v,w,x,y,z}
\forcsvlist{\MyMapTemplatePrefix {\def} {\widehat} {wh}} {A,B,C,D,E,F,G,H,I,J,K,L,M,N,O,P,Q,R,S,T,U,V,W,X,Y,Z}
\forcsvlist{\MyMapTemplatePrefix {\def} {\widehat} {wh}} {a,b,c,d,e,f,g,h,i,j,k,l,m,n,o,p,q,r,s,t,u,v,w,x,y,z}
\forcsvlist{\MyMapTemplatePrefixNew {\def} {\widehat}{\mathbf} {hb}} {A,B,C,D,E,F,G,H,I,J,K,L,M,N,O,P,Q,R,S,T,U,V,W,X,Y,Z}
\forcsvlist{\MyMapTemplatePrefixNew {\def} {\widehat}{\mathbf} {hb}} {a,b,c,d,e,f,g,h,i,j,k,l,m,n,o,p,q,r,s,t,u,v,w,x,y,z}
\forcsvlist{\MyMapTemplatePrefixNew {\def} {\overline}{\mathbf} {lb}} {A,B,C,D,E,F,G,H,I,J,K,L,M,N,O,P,Q,R,S,T,U,V,W,X,Y,Z}
\forcsvlist{\MyMapTemplatePrefixNew {\def} {\overline}{\mathbf} {lb}} {a,b,c,d,e,f,g,h,i,j,k,l,m,n,o,p,q,r,s,t,u,v,w,x,y,z}
\forcsvlist{\MyMapTemplatePrefix {\def} {\mathcal}{mc}} {A,B,C,D,E,F,G,H,I,J,K,L,M,N,O,P,Q,R,S,T,U,V,W,X,Y,Z}
\forcsvlist{\MyMapTemplatePrefix {\def} {\mathbb} {mb}} {A,B,C,D,E,F,G,H,I,J,K,L,M,N,O,P,Q,R,S,T,U,V,W,X,Y,Z}
\begin{document}

\title{Edit Temporal-Consistent Videos with Image Diffusion Model}

\author{Yuanzhi~Wang,
	    Yong~Li,
	    Xiaoya~Zhang,
	    Xin~Liu,
	    Anbo~Dai,
	    Antoni~B.~Chan,
	    Zhen~Cui
        % <-this % stops a space
\thanks{
	Yuanzhi Wang, Yong Li, Xiaoya Zhang, and Zhen Cui are with the PCA Lab, Key Laboratory of Intelligent Perception and Systems for High-Dimensional Information of Ministry of Education, School of Computer Science and Engineering, Nanjing University of Science and Technology, Nanjing, 210094, China.
	
	Xin Liu and Anbo Dai are with SeetaCloud, Nanjing, 210094, China.
	
	Antoni B. Chan is with the Department of Computer Science, City University of Hong Kong, Kowloon Tong, Hong Kong.
}% <-this % stops a space
}

% The paper headers
\markboth{Journal of \LaTeX\ Class Files,~Vol.~14, No.~8, August~2021}%
{Shell \MakeLowercase{\textit{et al.}}: A Sample Article Using IEEEtran.cls for IEEE Journals}

%\IEEEpubid{0000--0000/00\$00.00~\copyright~2021 IEEE}
% Remember, if you use this you must call \IEEEpubidadjcol in the second
% column for its text to clear the IEEEpubid mark.

\maketitle

\begin{abstract}
Large-scale text-to-image (T2I) diffusion models have been extended for text-guided video editing, yielding impressive zero-shot video editing performance. Nonetheless, the generated videos usually show spatial irregularities and temporal inconsistencies as the temporal characteristics of videos have not been faithfully modeled.
In this paper, we propose an elegant yet effective \textit{Temporal-Consistent Video Editing (TCVE)} method to mitigate the temporal inconsistency challenge for robust text-guided video editing.
In addition to the utilization of a pretrained T2I 2D Unet for spatial content manipulation, we establish a dedicated temporal Unet architecture to faithfully capture the temporal coherence of the input video sequences.
Furthermore, to establish coherence and interrelation between the spatial-focused and temporal-focused components, a cohesive spatial-temporal modeling unit is formulated. This unit effectively interconnects the temporal Unet with the pretrained 2D Unet, thereby enhancing the temporal consistency of the generated videos while preserving the capacity for video content manipulation.
Quantitative experimental results and visualization results demonstrate that TCVE achieves state-of-the-art performance in both video temporal consistency and video editing capability, surpassing existing benchmarks in the field. 
\end{abstract}


\section{Introduction}

Recently, diffusion-based generative models~\cite{ddpm,scoresde,imder,DiffFashion} have shown remarkable image~\cite{stablediffusion,imagen,controlnet,Dreambooth} and video~\cite{Imagenvideo,Make-A-Video,Alignyourlatents,Animatediff} generation capabilities via diverse text prompts.
It brings the large possibility to edit real-world visual content by merely editing the text prompts.

\begin{figure}[htb]
	\centering{\includegraphics[width=\linewidth]{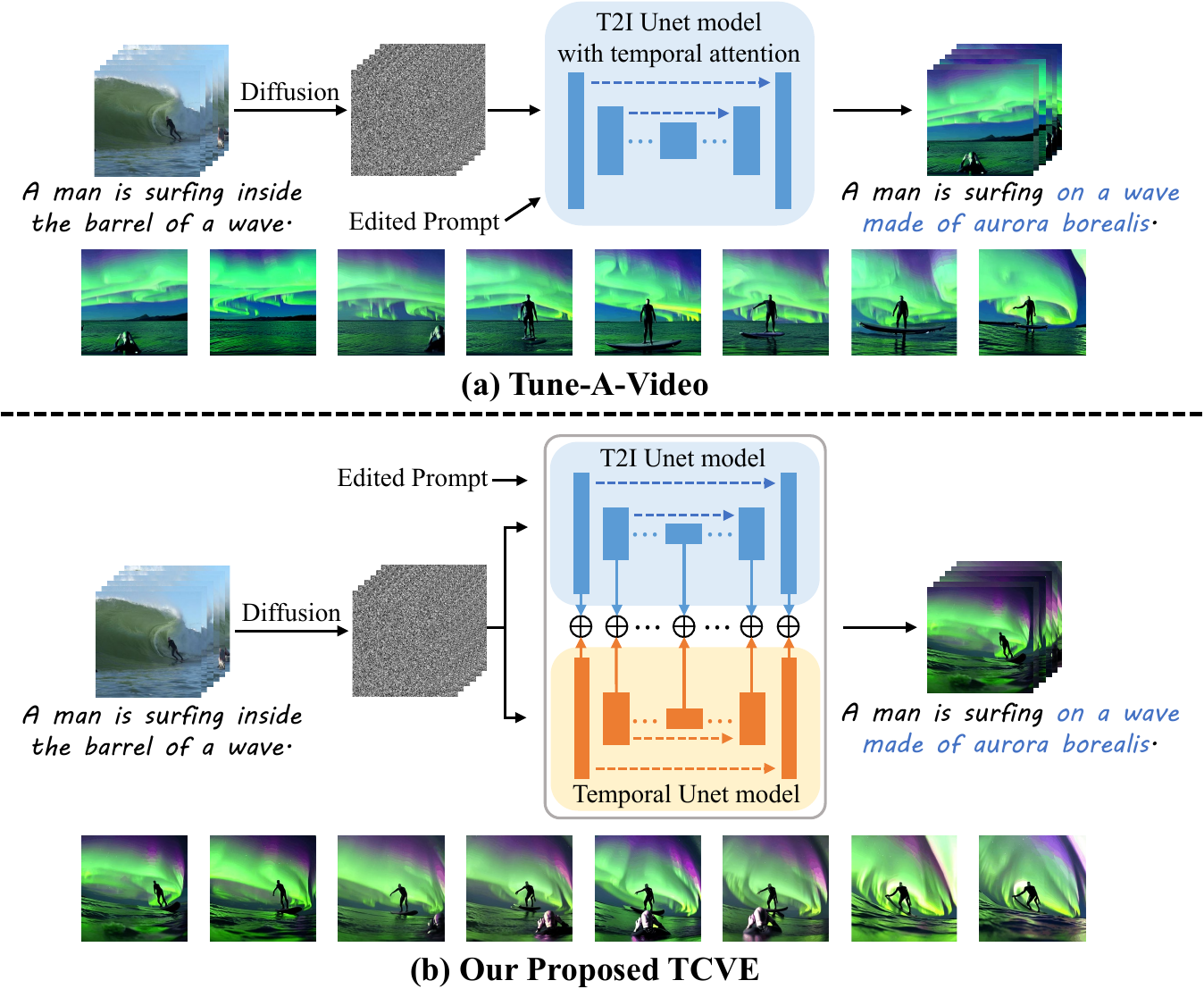}}
	\caption{Two frameworks for text-guided video editing. (a) shows the Tune-A-Video~\cite{TAV} method. This pioneer work suffers from flickering artifacts and the surfing stance of people is almost distorted.  (b) illustrates our proposed Temporal-Consistent Video Editing (TCVE) method. TCVE exploits a dedicated temporal Unet to preserve the temporal consistency. As a comparison, TCVE faithfully manipulates image content in accordance with the provided prompt and shows encouraging temporal coherency.
	}
	\label{fig:1}
\end{figure}

Based on the publicly available large-scale pretrained text-to-image (T2I) models, e.g., Stable Diffusion~\cite{stablediffusion}, researchers have developed various text-guided diffusion-based image editing methods~\cite{Blendeddiffusion,Plug-and-play}.
To edit images, the main idea is to leverage deterministic DDIM~\cite{DDIM} for the image-to-noise inversion, and then the inverted noise is gradually denoised to the edited images under the condition of the edited prompt.

When it comes to text-guided video editing, a seemingly direct approach involves an extension of the aforementioned paradigm to encompass video content. Nevertheless, this paradigm is riddled with two formidable challenges: firstly, the absence of readily accessible large-scale pretrained text-to-video (T2V) diffusion models; and secondly, the typically resource-intensive nature of training or refining T2V models for video editing purposes. Consequently, an approach grounded in text-to-image (T2I) models appears to hold greater potential value compared to one centered on video, primarily owing to the plethora of open-source T2I models available within the broader community.

Some researchers have exploited the pretrained T2I models for  text-guided video editing, e.g., Tune-A-Video~\cite{TAV} flattens the temporal dimensionality of the source video and then manipulates the spatial content frame-by-frame using the T2I model to generate the target video, as shown in Fig.~\ref{fig:1} (a).
In this case, the additional temporal attention modules are incorporated into the T2I model, while the source video and the corresponding prompt are used to train these temporal attention modules to preserve the temporal consistency among frames.
Subsequently, Qi \textit{et al.}~\cite{fatezero} designed a fusing attention mechanism based on Tune-A-Video to fuse the attention maps from the inversion and generation process to preserve the motion and structure consistency.
As verified in Fig.~\ref{fig:1} (a), these previous methods still suffer from flickering artifacts and inconsistency among consecutive frames due to incomplete and inconsistent temporal modeling.
For the above video-editing paradigms, the temporal attention modules are directly injected into each stage of the spatial-only T2I Unet model for temporal modeling.
This means that the input of the temporal attention module is merely spatial-aware and temporal modeling capability might not be reliable or faithfully.

In this paper, we aim to challenge the above limitations by proposing an elegant yet effective Temporal-Consistent Video Editing (TCVE) method, as shown in Fig.~\ref{fig:1} (b).
To model the temporal coherency, we construct a temporal Unet model to facilitate temporal-focused modeling.
In the temporal Unet model, each residual block is built by stacked temporal convolutional layers. The input video-based tensor is reshaped into a temporal-focused manner for reliable temporal modeling.
In particular, to bridge the temporal Unet and the pretrained T2I 2D Unet, we establish a spatial-temporal modeling unit to consolidate the temporal consistency while maintaining the video editing capability.
In contrast to prior work, TCVE can faithfully mitigate the flickering artifacts between consecutive frames, as shown in the results of Fig.~\ref{fig:1} (b).
In summary, the contributions of this work can be concluded as:
\begin{itemize}
	\item To mitigate the temporal inconsistency issue for reliable text-guided video editing, we present a well-designed and efficient Temporal-Consistent Video Editing (TCVE) method. TCVE strategically employs a dedicated temporal Unet model to guarantee comprehensive and coherent temporal modeling.
	\item To bridge the temporal Unet and pretrained T2I 2D Unet, we introduce a cohesive spatial-temporal modeling unit. This unit is adept at capturing both temporal and spatial information, thereby enhancing the temporal consistency of the edited video while concurrently preserving the capacity for video editing.
	\item We perform extensive experiments on text-guided video editing datasets and achieve superior or comparable results. Quantitative and visualization results demonstrate that the flickering artifacts and temporal inconsistency are effectively mitigated.
\end{itemize}

\section{Related Works}

\subsection{Text-to-image Generation}

Text-to-image (T2I) generation task aims to generate photorealistic images that semantically match given text prompts~\cite{AlignDRAW,t2i-icml,ControllableT2I,DALL-E,t2i-tmm1,t2i-tmm2}.
The main idea of T2I generation is to utilize the current generative modeling paradigms such as Generative Adversarial Networks (GANs)~\cite{GANs,gansurvvey}, normalizing flows~\cite{Glow,dicmor}, diffusion models~\cite{ddpm,scoresde}, to construct a conditional generative model conditioned on text embedding.

The pioneer work of T2I generation is AlignDRAW~\cite{AlignDRAW}, which generated images from natural language descriptions by applying sequential deep learning techniques to conditional probabilistic models.
Subsequently, Reed \textit{et al.}~\cite{t2i-icml} proposed a Text-conditional GAN model that is the first end-to-end differential architecture from the word-level to pixel-level.
Further, some researchers have developed several autoregressive methods to exploit large-scale text-image data for T2I generation, such as DALL-E~\cite{DALL-E} and Parti~\cite{Parti}.
Recently, due to the powerful capability of estimating data distribution and the stable training process, diffusion-based generative models have achieved unprecedented success in the T2I generation domain~\cite{DALLE-2,imagen,stablediffusion,GLIDE,controlnet,Reco}.
For example, Ramesh \textit{et al.}~\cite{DALLE-2} proposed the DALLE-2 that uses CLIP-based~\cite{clip} feature embedding to build a T2I diffusion model with improved text-image alignments.
Saharia \textit{et al.}~\cite{imagen} designed robust cascaded diffusion models for high-quality T2I generation.
Rombach \textit{et al.}~\cite{stablediffusion} proposed a novel Latent Diffusion Model (LDM) paradigm that projects the original image space into the latent space of an autoencoder to improve training efficiency.
Benefiting from the excellent generation quality and efficiency of LDM, it has become the most popular T2I generation model.
Therefore, researchers have explored and exploited LDM paradigm to develop various works~\cite{controlnet,Reco,Dreambooth,Spatext}.
For instance, Zhang \textit{et al.}~\cite{controlnet} proposed a ControlNet that appended additional conditions, such as Canny edges, depth maps, human poses, to provide diverse generative capabilities.
Reco \textit{et al.}~\cite{Reco} designed a region-controlled T2I diffusion model based on pretrained Stable Diffusion to achieve controlled generation.
Considering the nice properties of LDM, we exploit it as a backbone model for text-guided video editing.

\subsection{Text-to-video Generation}
Despite major advances in T2I generation, text-to-video (T2V) generation is still lagging behind due to the lack of large-scale text-video datasets and the thousands of times harder to train compared to T2I diffusion models.
To achieve the T2V diffusion models, some researchers have attempted to propose various methods~\cite{VDM,Magicvideo,Imagenvideo,Alignyourlatents,Make-A-Video,higen,Modelscope,Pvdm}.
For instance, \cite{VDM} proposed a Video Diffusion Model (VDM) that is a naive extension of the standard image diffusion models, and the original 2D Unet was replaced by space-only 3D Unet to fit the video samples.
Subsequently, Ho \textit{et al.}~\cite{Imagenvideo} combined VDM with Imagen~\cite{imagen} and designed an Imagen Video to generate high-definition videos.
Blattmann \textit{et al.}~\cite{Alignyourlatents} applied the LDM paradigm to high-resolution video generation, called Video LDM.
In addition to this, some works aim to utilize the pretrained T2I diffusion models to conduct T2V generation, which can mitigate the difficulty of training from scratch~\cite{Animatediff,Text2video-zero,DSNET}.
For example, Guo \textit{et al.}~\cite{Animatediff} proposed the Animatediff that inserts the motion (i.e., temporal) modules into pretrained T2I 2D Unet to facilitate T2V generation.
Although these T2V methods are capable of generating high-quality videos, there are still many issues that hinder the development of this field, such as large-scale privatized training data, non-public well-trained models, and high training costs, etc.

\begin{figure*}[t]
	\centering{\includegraphics[width=0.97\linewidth]{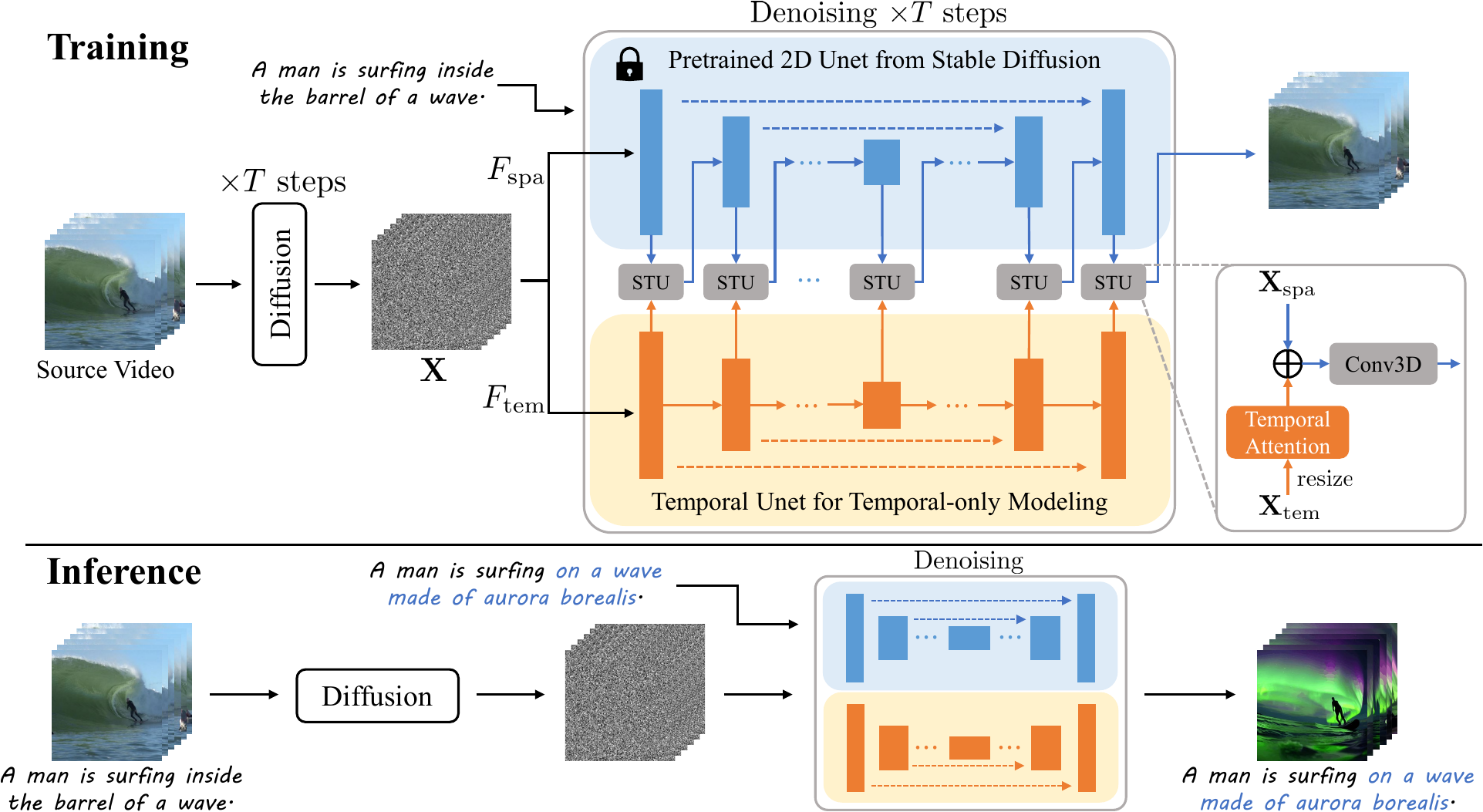}}
	\caption{The framework of TCVE. Given a text-video pair as input, TCVE leverages the \textit{pretrained 2D Unet from Stable Diffusion}~\cite{stablediffusion} and our proposed \textit{temporal Unet} for text-guided video editing. 
		The input video is first diffused into a noisy video $\X \in \mathbb{R}^{b\times c\times f\times h\times w}$, where $b, c, f, h, w$ denote batch, channel, frame, height, and width dimensionality, respectively. Then, $\X$ is reshaped into a spatial-dominated tensor (i.e., $F_{\text{spa}}(\X)\in \mathbb{R}^{(b\times f)\times c\times h\times w}$) and a temporal-dominated tensor (i.e., $F_\text{tem}(\X) \in \mathbb{R}^{(b\times h\times w)\times c\times f}$) for subsequent spatial and temporal focused modeling. 
		To bridge the temporal Unet and pretrained 2D Unet, we establish a Spatial-Temporal modeling Unit (STU) that adaptively fuses the spatial- and temporal-aware feature ($\X_{\text{spa}}$,  $\X{_\text{tem}}$). During training, we update the parameters of temporal Unet and STUs with the standard diffusion training loss. For inference, we generate a new video from the source video under the guidance of a modified prompt.
	}
	\label{fig:2}
\end{figure*}

\subsection{Text-guided Video Editing}
Recent text-guided diffusion-based image editing methods~\cite{SDEdit,Blendeddiffusion,Plug-and-play,DiffEdit,Imagic} achieve promising image editing results.
Despite the great success, text-guided video editing is still lagging behind, as it faces the same difficulties as the development of the T2V models.
Some works attempt to challenge this problem~\cite{text2live,gen1,loeschcke2022text,TAV,fatezero,stablevideo}.
For example, Text2Live~\cite{text2live} and StableVideo~\cite{stablevideo} allowed some texture-based video editing with edited prompts.
However, Text2Live and StableVideo depend on the layered neural atlases~\cite{atlases}, thus the editing capabilities are often limited.
Dreamix~\cite{dreamix} and Gen-1~\cite{gen1} aimed to utilize VDM to conduct video editing, but training VDM required large-scale datasets and tremendous computational resources.
Moreover, their training data and pretrained models are not publicly available.
Recently, some works have exploited the pretrained T2I diffusion models to conduct efficient text-guided video editing with a single GPU device~\cite{TAV,fatezero}.
The first work is Tune-A-Video~\cite{TAV}, which flattens the temporal dimensionality of the source video and then edits it frame-by-frame using the T2I diffusion model to generate the target video.
Of these, the extra temporal attention modules are incorporated into the T2I diffusion model to preserve the temporal consistency among frames.
FateZero~\cite{fatezero} then improved the Tune-A-Video by designing a fusing attention mechanism to preserve the motion and structure.
However, these previous methods still result in flickering artifacts and inconsistency among frames due to incomplete and inconsistent temporal modeling caused by alternating spatial and temporal modeling.
Our proposed TCVE has an essential difference from these prior works because we design a temporal Unet as an independent temporal branch, which could guarantee complete temporal awareness.

\section{Preliminaries}

\subsection{Latent Diffusion Models}
As one of the most popular diffusion-based generative paradigms, Latent Diffusion Models (LDMs)~\cite{stablediffusion} was proposed to diffuse and denoise the latent space of an autoencoder to improve training efficiency.
Specially, an encoder $\mcE$ projects original image $x$ into a low-resolution latent state $z=\mcE(x)$, and $z$ can be reconstructed back to the original image $x\approx\mcD(z)$ by a decoder $\mcD$.
Then, a denoising Unet $\epsilon_{\theta}$ with cross-attention and self-attention~\cite{attention} is trained to denoise Gaussian noise into clean latent state $z$ using the following objective:
\begin{equation}
\label{ldmloss}
\mcL_{\text{LDM}} = \mathbb{E}_{z_{0},\epsilon\sim\mcN(\0,\I),t\sim \mcU(1, T)} [ \|\epsilon-\epsilon_{\theta}(z_{t}, t, p)\|_2^2],
\end{equation}
where $p$ is the conditional text prompt embedding that is often extracted from the CLIP text encoder~\cite{clip}.
$z_{t}$ is a diffused sample at timestep $t$, $\mcN$ is a Gaussian distribution, and $\mcU$ is a Uniform distribution.

\subsection{DDIM Sampler and Inversion}
During inference, DDIM~\cite{DDIM} sampler was employed to convert a Gaussian noise $z_{T}$ to a clean latent state $z_{0}$ in a sequence of timestep $t=T\rightarrow1$ with the following iteration rule $\text{DDIM}_\text{smp}:{z}_{t}\xrightarrow[]{\epsilon_\theta} {z}_{t-1}$,
\begin{equation}
\label{iteration rule}
z_{t-1} = \sqrt{\alpha_{t-1}}\frac{z_{t}-\sqrt{1-\alpha_{t}}\epsilon_{\theta}}{\sqrt{\alpha_{t}}} + \sqrt{1-\alpha_{t-1}}\epsilon_{\theta},
\end{equation}
where $\alpha_{t}$ is a noise scheduling parameter defined by~\cite{ddpm}.
Next, the DDIM inversion was proposed to project a clean latent state $z_{0}$ into a noisy latent state $\hat{z}_{T}$ in a sequence of revered timestep $t=1\rightarrow T$ with the following iteration rule $\text{DDIM}_\text{inv}:\hat{z}_{t-1}\xrightarrow[]{\epsilon_\theta} \hat{z}_{t}$,
\begin{equation}
\label{revered iteration rule}
\hat{z}_{t} = \sqrt{\alpha_{t}}\frac{\hat{z}_{t-1}-\sqrt{1-\alpha_{t-1}}\epsilon_{\theta}}{\sqrt{\alpha_{t-1}}} + \sqrt{1-\alpha_{t}}\epsilon_{\theta}.
\end{equation}
Intuitively, $\hat{z}_{T}$ can be denoised into a clean latent state $\hat{z}_{0} = \text{DDIM}_\text{smp}(\hat{z}_{T}, p) \approx z_{0}$ with a classifier-free guidance whose scale factor set as 1.
Current image editing methods~\cite{p2p,Nulltext} use a large classifier-free guidance scale factor ($\gg 1$) to edit the latent with an edited prompt $p_{\text{edit}}$ as $\hat{z}_{0}^{\text{edit}}=\text{DDIM}_\text{smp}(\hat{z}_{T}, p_{\text{edit}})$, and then use decoder $\mcD$ to map  $\hat{z}_{0}^{\text{edit}}$ into an edited image $x_{\text{edit}}$.

\section{Method}

\subsection{Problem Formulation}
Let $\mcV=(v_1,v_2,\cdots, v_m)$ denotes a source video that contains $m$ video frames. $p_{\text{sour}}$ and $p_{\text{edit}}$ denote the the source prompt describing $\mcV$ and the edited target prompt, respectively.
The goal of text-guided video editing is to generate a new video
$\mcV_{\text{edit}}$ from source video $\mcV$ under the condition of the edited prompt $p_{\text{edit}}$.
For example, consider a video and a source prompt \textit{``A man is surfing inside the barrel of a wave''}, and assume that the user wants to change the background of wave while preserving the motion.
The user can directly modify the source prompt such as \textit{``A man is surfing on a wave made of aurora borealis''}.
Recent excellent works, e.g., Tune-A-Video~\cite{TAV}, exploited the pretrained T2I diffusion models to conduct video editing tasks.
However, they mostly emphasize spatial content generation, although the temporal attention modules are also used to facilitate temporal awareness.

Our main idea is to build an independent temporal diffusion network through using temporal convolutional layers to model the temporal information of videos based on the T2I Unet model, as shown in the upper part in Fig.~\ref{fig:2}.
In addition to the utilization of a pretrained 2D Unet for spatial content manipulation, we establish a dedicated temporal Unet to faithfully capture the temporal coherence of the input video.

Concretely, the input video $\mcV$ is first encoded by $\mcE$ and inverted to the noise by DDIM inversion.
Then, the inverted noise is gradually denoised to the edited video frames through DDIM sampler under the edited prompt $p_{\text{edit}}$ and decoded by $\mcD$.
Among them, an input video tensor is flattened into a \textit{spatial}-dominated tensor via the \textit{pretrained 2D Unet} and a \textit{temporal}-dominated tensor via the \textit{temporal UNet}, respectively.
Then, the spatial-/temporal-dominated tensors are separately injected into 2D and temporal Unet to enhance the spatial and temporal awareness.
Formally, the generation process of edited video frames is abstractly defined as:
\begin{align}
\label{edit}
\mcV_{\text{edit}} = \mcD\left[\!\!\!
\begin{array}{c}
\text{DDIM}_{\text{smp}}(\text{DDIM}_{\text{inv}}(F_\text{spa}(\mcE(\mcV)),\theta_\text{spa}),p_{\text{edit}},\theta_\text{spa})\\
\text{DDIM}_{\text{smp}}(\text{DDIM}_{\text{inv}}(F_\text{tem}(\mcE(\mcV)),\theta_\text{tem}),p_{\text{edit}},\theta_\text{tem})
\end{array}
\!\!\!\right],
\end{align}
where $F_\text{spa}$ and $F_\text{tem}$ denote the flattening operations used to generate the spatial-/temporal-dominated tensors, as shown in Fig.~\ref{fig:2}.
$\theta_\text{sps}$ and $\theta_\text{tem}$ denote the parameters of T2I 2D Unet and temporal model, respectively.

\subsection{Network Architecture}
We now illustrate the proposed Temporal-Consistent Video Editing (TCVE) network architecture, as shown in Fig.~\ref{fig:2}.
The network architecture is mainly composed of three parts: pretrained \textbf{T2I 2D Unet}, \textbf{Temporal Unet}, and \textbf{Spatial-Temporal modeling Unit}. Below, we describe these modules in detail.

\textbf{T2I 2D Unet.}
The common T2I diffusion model such as Stable Diffusion~\cite{stablediffusion}) typically consists of a 2D  spatial-only Unet model~\cite{unet}, which is a neural network based on a spatial downsampling pass followed by an upsampling pass with skip connections.
In such 2D Unet architecture, several 2D convolutional residual blocks and transformer blocks are stacked to encode the spatial information.
Each transformer block is mainly composed of a spatial self-attention layer that leverages pixel locations to capture spatial dependency and a cross-attention layer to capture correlations between embedded image feature and embedded prompt feature. The latter cross-attention layer is the core of condition generation, e.g., text prompt.
Intuitively, the original 2D Unet model cannot well encode continuous temporal variation information due to the lack of dynamic sequence modeling. Hence, the generated videos from T2I without expert sequence models would often result in flickering artifacts. To suppress those artifacts effectively, we specifically design a temporal diffusion model to compensate for the generated content information, which is introduced in the next parts.

\begin{figure*}[htb]
	\centering{\includegraphics[width=\linewidth]{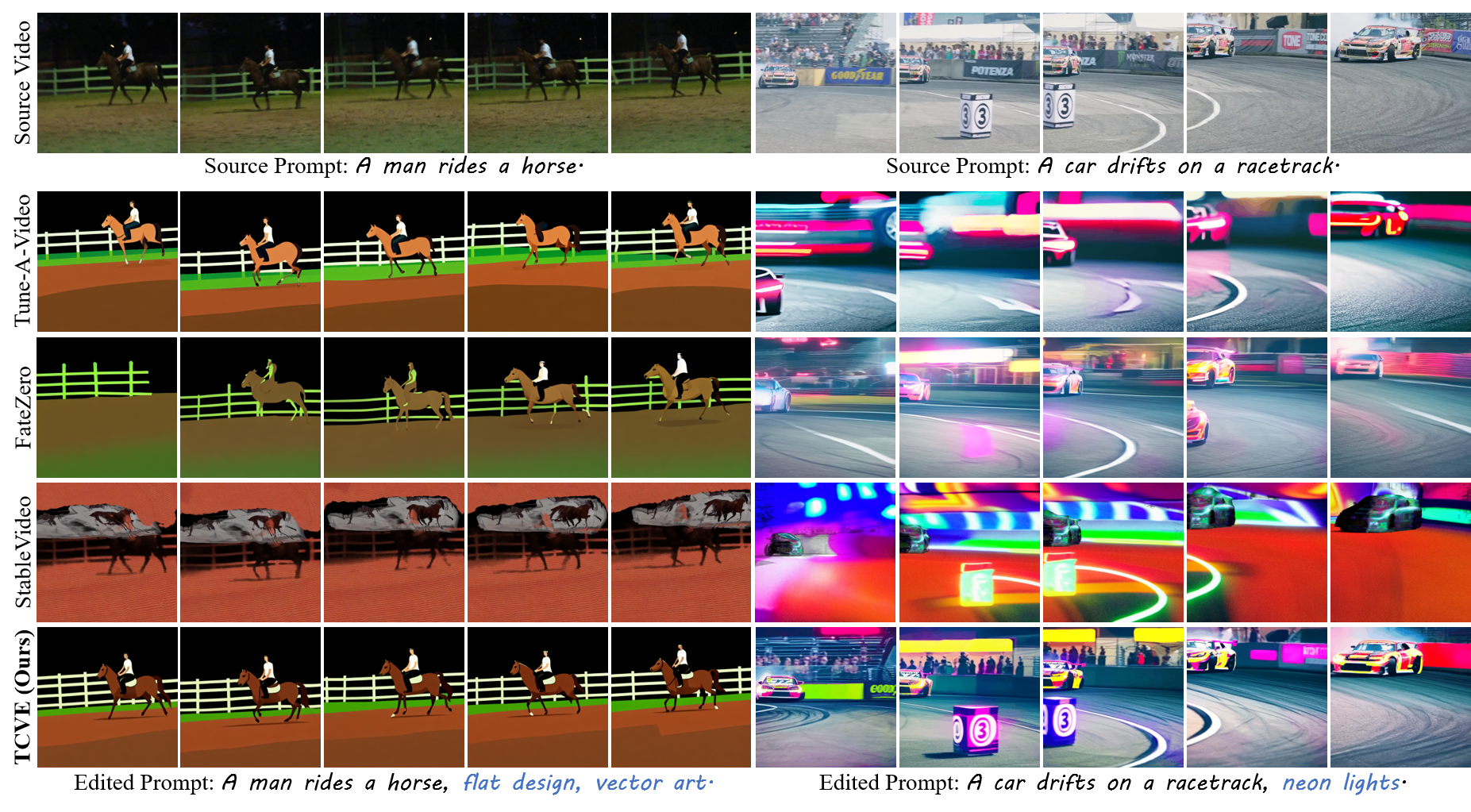}}
	\caption{Visualization of style transfer. Tune-A-Video and FateZero may cause some frame inconsistency, and StableVideo tends to disrupt the texture information of foreground objects (i.e., horse and car). In contrast, our TCVE generates temporally consistent videos while effectively transferring style.
	}
	\label{style transfer}
\end{figure*}

\textbf{Temporal Unet.} 
As shown in Fig.~\ref{fig:2}, we design a temporal model to reliably model the temporal consistency. To well-align the spatial T2I Unet, we also choose the Unet architecture as the temporal model, whereas operating on temporal axis with downsampling pass followed by an upsampling pass with skip connections.
Different from the 2D spatial Unet model, the temporal Unet is composed of stacked temporal (i.e., 1D) convolutional residual blocks.
Considering an input video-based tensor $\X \in \mathbb{R}^{b\times c\times f\times h\times w}$, where $b, c, f, h, w$ indicate batch size, channel number, frame number, height, and width, respectively.
The spatial dimensions $h$ and $w$ will first be reshaped to the batch dimension, resulting in $b\times h\times w$ sequences at the length of $f$, i.e., $F_\text{tem}(\X) \in \mathbb{R}^{(b\times h\times w)\times c\times f}$.
Then, the reshaped tensor is injected into temporal convolutional residual blocks for temporal axis downsampling and upsampling.
Take a downsampling stage as an example, the input tensor size is $(b\times h\times w)\times c\times f$ and the output tensor size is $(b\times h\times w)\times 2c\times \frac{f}{2}$, and the upsampling is vice versa.
Intuitively, the temporal Unet can completely and consistently model temporal information due to the temporal-aware input and output for each block.

\textbf{Spatial-Temporal modeling Unit (STU).}
Another question is how to connect the temporal Unet model and the 2D Unet model.
To bridge the above two models, we design a Spatial-Temporal modeling Unit (STU) to perceive both temporal and spatial information.
As shown in Fig.~\ref{fig:2}, the STU is mainly composed of a temporal attention block and a 3D convolutional block. After performing the spatial-/temporal-focused modeling, we are supposed to obtain the spatial-aware feature $\X_{\text{spa}}$ and temporal-aware feature $\X_{\text{tem}}$, respectively. The STU takes $\X_{\text{spa}}$ and $\X_{\text{tem}}$ as input. In order to facilitate subsequent feature fusion, $\X_{\text{tem}}$ is aligned to the size of $\X_{\text{spa}}$ by resizing the same shape. Then, in particular, a temporal attention block is used to enhance temporal awareness of the resized $\X_{\text{tem}}$, formulated as:
\begin{equation}
\X^{\text{att}}_{\text{tem}}= \text{Attention}(\Q,\K,\V)= \text{softmax}(\frac{\Q\K^{\top}}{\sqrt{d}})\V,
\end{equation}
where $\Q=\W_{q} \X_{\text{tem}}$, $\K=\W_{k} \X_{\text{tem}}$, and $\V=\W_{v} \X_{\text{tem}}$, and $\W_q, \W_k, \W_v$ are the learnable parameters.
This attention operation enables the module to capture the temporal dependencies between features at the same spatial location across the temporal axis.
After that, $\X_{\text{spa}}$ and $\X^{\text{att}}_{\text{tem}}$ are fused in a weighted manner: $\X_{\text{fuse}}=\X_{\text{spa}}+\lambda \X^{\text{att}}_{\text{tem}}$, where $\lambda=0.1$ is the balance factor.
Finally, a 3D convolutional block is utilized to conduct $\X_{\text{fuse}}$ for spatial-temporal modeling due to its nice property for processing video-based context, thereby improving the temporal consistency of the generated video while maintaining the editing capability.

\subsection{Training and Inference}
The paradigms of training and inference are shown in Fig.~\ref{fig:2}.
During the training period, the source video and source prompt are used to train the temporal Unet model and STUs with the original LDM objective in Eq.~\ref{ldmloss}, and the parameters of the pretrained T2I 2D Unet model (we use Stable Diffusion in this work) are frozen and not trainable.
Note that the training period does not need to train for each edited prompt individually, thus our method is a zero-shot video editing paradigm.
During the inference period, we use the way defined by Eq.~\ref{edit} to edit the target video.
Our experiments demonstrate such training and inference strategy is effective in accurately delivering the motion and structure from the source video to the edited videos.

\section{Experiments}

\subsection{Implementation Details}
\begin{figure*}[htb]
	\centering{\includegraphics[width=1.0\linewidth]{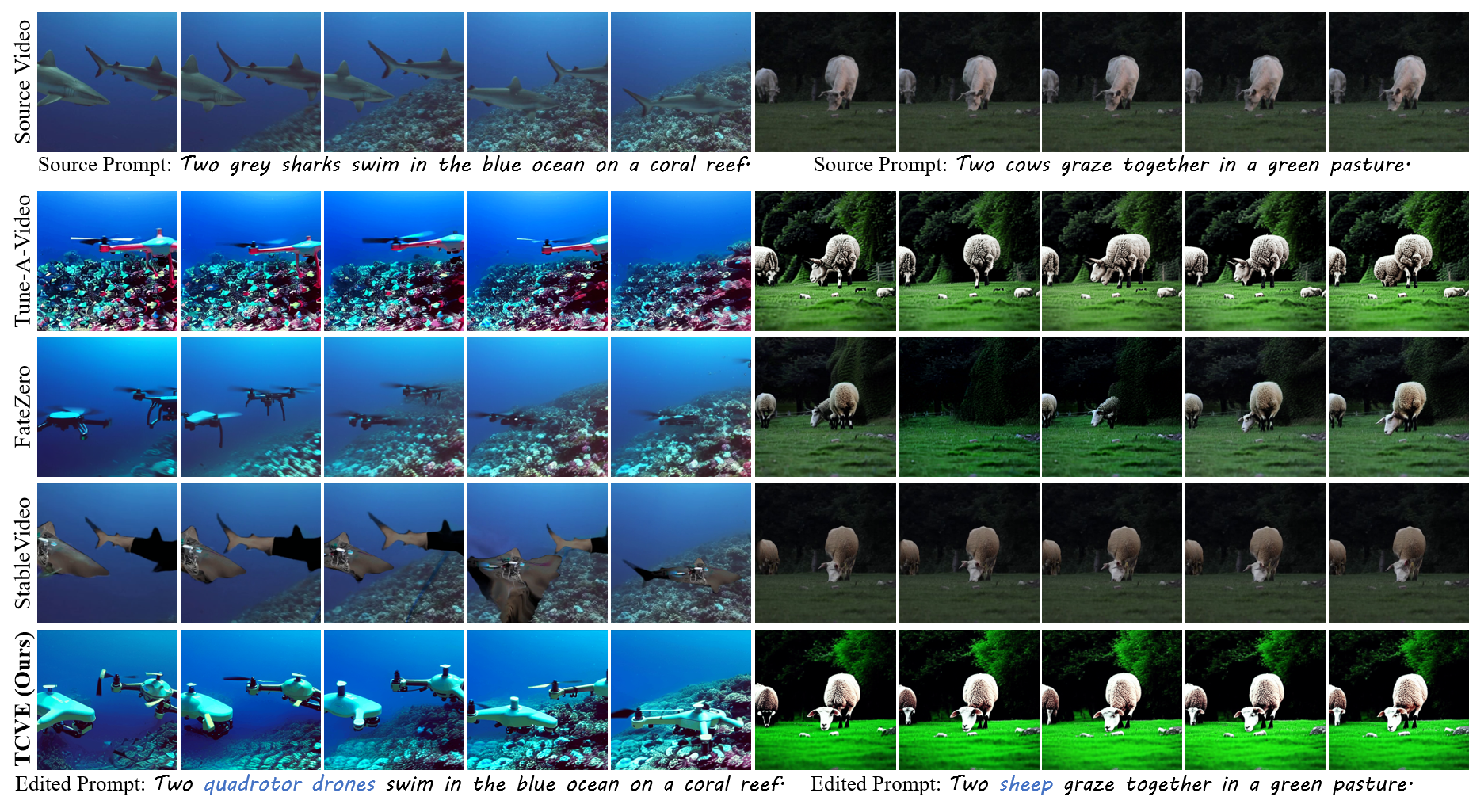}}
	\caption{Qualitative comparison of object editing. Intuitively, our proposed TCVE can faithfully alter the objects according to the edited prompts while maintaining other video attributes.
	}
	\label{object editing}
\end{figure*}

\begin{figure*}[htb]
	\centering{\includegraphics[width=1.0\linewidth]{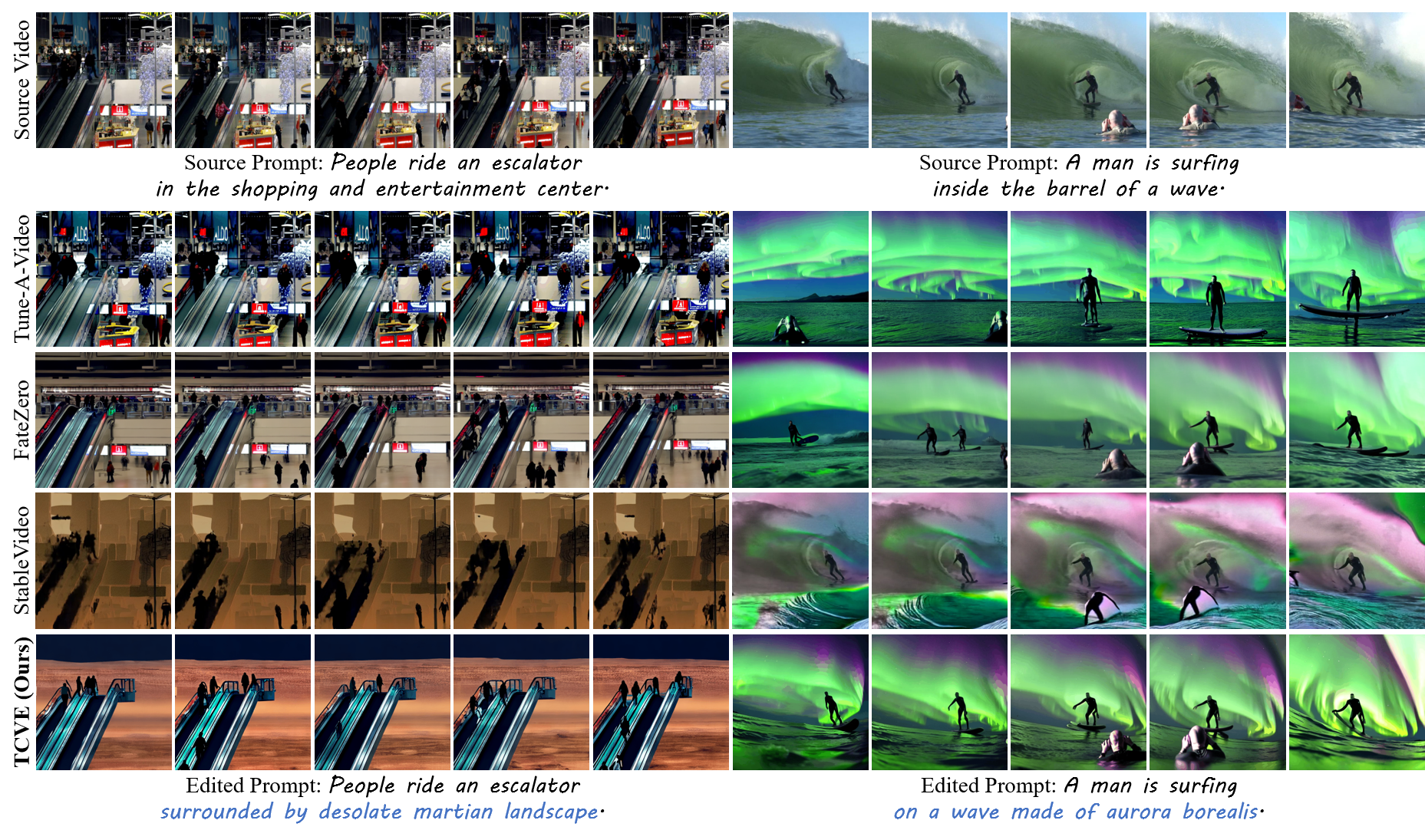}}
	\caption{Qualitative comparison of background change. In contrast to these prior works, TCVE can faithfully manipulate the background and preserve the smoothness between frames. 
	}
	\label{background change}
\end{figure*}

\begin{figure*}[htb]
	\centering{\includegraphics[width=1.0\linewidth]{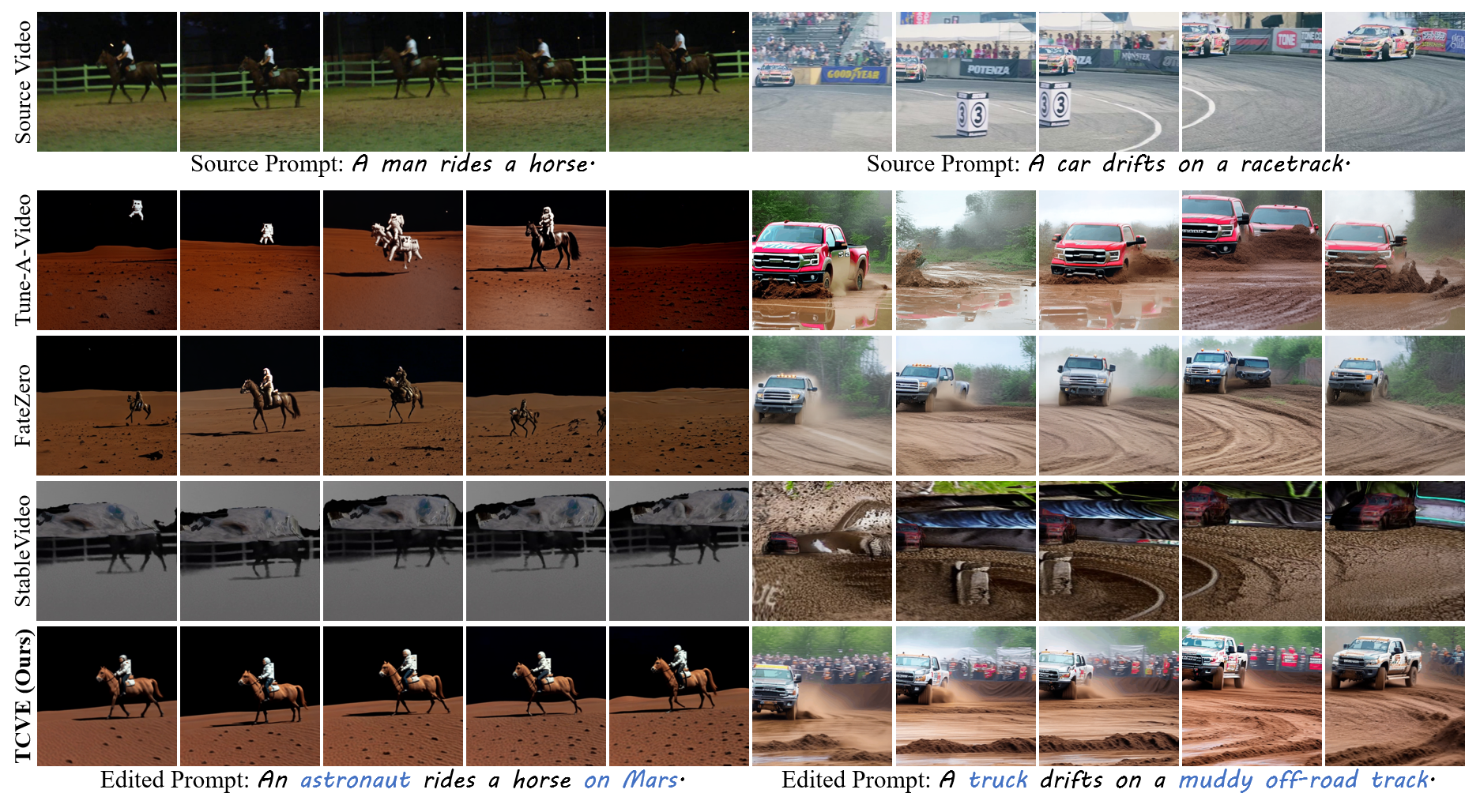}}
	\caption{Qualitative comparison of multiple-object editing. TCVE can generate temporally-coherent videos, and both background and objects are well edited.
	}
	\label{multiple}
\end{figure*}

Our TCVE is based on public pretrained Stable Diffusion v1.4\footnote{https://huggingface.co/CompVis/stable-diffusion-v1-4}.
We conduct experiments on several videos from the latest text-guided video editing dataset LOVEU-TGVE-2023\footnote{https://github.com/showlab/loveu-tgve-2023} and the video samples used in~\cite{stablevideo}.
Each video has 4 different edited prompts for 4 applications: style transfer, object editing, background change, and multiple-object editing.
\textbf{Style transfer} aims to transfer videos into a variety of styles.
For example, we can transfer a real-world video into a vector art style, as shown in Fig.~\ref{style transfer}.
\textbf{Object editing} allows the users to edit the objects of the video.
As shown in Fig.~\ref{object editing}, we can replace ``Two gray sharks'' with ``Two quadrotor drones''.
\textbf{Background change} can enable users to change the video background, i.e., the place where the object is, while preserving the consistency in the movement of the object.
For example, we can change the background of the ``shopping and entertainment center'' to the ``martian landscape'', as shown in Fig.~\ref{background change}.
\textbf{Multiple-object editing} aims to edit multiple contents, e.g., perform object editing and background change, as shown in Fig.~\ref{multiple}.
We utilize Pytorch to implement all experiments with a RTX 3090 GPU.
In the training stage, the models are trained by Adam optimizer with learning rate $3\times 10^{-5}$, and each video is fixed for 100 iterations.
During inference, we use DDIM inversion and DDIM sampler~\cite{DDIM} with 50 steps and classifier-free guidance with a guidance scale of 12.5 in our experiments.

\textbf{Evaluation Metrics.}
We consider three evaluation metrics that are used by the latest text-guided video editing dataset LOVEU-TGVE-2023 to measure the quality of generated videos.
\textbf{Frame Consistency} is to measure the temporal consistency
in frames by computing CLIP image embeddings on all frames of output video and report the average cosine similarity between all pairs of video frames.
\textbf{Textual Alignment} is to measure textual faithfulness of edited video by computing the average CLIP score between all frames of output video and corresponding edited prompt.
\textbf{PickScore}~\cite{pickscore} is to measure human preference for text-to-image generation models.
We compute the average PickScore between all frames of the output video and the corresponding edited prompt.

\subsection{Baseline Comparisons}
We compare our method with the three latest baselines: 1) Tune-A-Video~\cite{TAV} is a pioneer in efficient text-guided video editing using pretrained T2I diffusion models.
2) FateZero~\cite{fatezero} is an improved method with the fusing attention mechanism based on Tune-A-Video.
3) StableVideo~\cite{stablevideo} is an atlas-based method that exploits the pretrained T2I diffusion models to edit 2D layered atlas images for text-guided video editing.
Below, we analyze quantitative and qualitative experiments.

\begin{table}[!ht]
	\centering
	\caption{Quantitative comparison with evaluated baselines.}\label{tab:1}
	\setlength{\tabcolsep}{2pt}
	\scalebox{1.0}{
		\begin{tabular}{c|c|c|c}
			\hline
			Methods & Frame Consistency & Textual Alignment &  PickScore \\
			\hline
			\hline
			Tune-A-Video~\cite{TAV} &91.99  &  26.52 & 20.29 \\
			FateZero~\cite{fatezero} & 92.64 & 25.88  & 20.31 \\
			StableVideo~\cite{stablevideo} & 93.46 & 24.05  & 19.34 \\ 
			TCVE (Ours) & \textbf{94.44} & \textbf{27.58} & \textbf{20.50} \\
			\hline
	\end{tabular}}
\end{table}

\subsubsection{Quantitative results}
Tab.~\ref{tab:1} lists the quantitative results of different methods.
From these results, we can observe that TCVE achieves the best video editing performance under three evaluation metrics.
In particular, TCVE gains considerable performance improvements in the Frame Consistency and Textual Alignment.
This occurrence can be ascribed to the exhaustive and uniform modeling of spatial-temporal information accomplished by TCVE. This accomplishment is facilitated by the deployment of an dedicated temporal Unet as well as STU, thereby significantly amplifying the temporal coherence of the generated video.
Further visualization analysis for the generated video will be provided in the next part.

\subsubsection{Qualitative results}
We provide some visual comparison of our TCVE against three baselines in four editing tasks. 

\textbf{Style transfer.} Fig.~\ref{style transfer} shows the visual results of style transfer. All methods seem to transfer the video style at first glance.
However, the results of three baselines suffer from frame inconsistency or texture distortions.
For Tune-A-Video w.r.t the left sample, the man disappears in the $4^{th}$ frame. 
For the right sample, the car has been distorted.
For FateZero w.r.t the left sample, the object disappears in the $1^{th}$ frame and the horse color shows inconsistency.
For the right sample, two cars appear in the $4^{th}$ frame.
For StableVideo, the texture information of foreground objects (i.e., horse and car) in two samples are severely distorted and disrupted, and the ``flat design, vector art'' style is also unsatisfactorily transferred in the left sample.
In contrast, TCVE can produce temporally smooth videos while successfully editing the video style.

\textbf{Object editing.} Fig.~\ref{object editing} exhibits the comparison of object editing task.
Tune-A-Video shows obvious limitations: the solitary drone is not consistent with the goal of two drones (for the left sample) and the sheep orientation have not been preserved (for the right sample).
For the FateZero, evident flickering and temporal inconsistency also exist, e.g.,  sudden distortion of the two drones (for the left sample).
StableVideo does not have the ability to edit and deform foreground objects due to the intrinsic limitations of atlas-based methods, i.e., restrictions on the foreground opacity values~\cite{atlases}. 
Compared with them, TCVE faithfully alters the object according to the prompt while maintaining other video attributes.

\textbf{Background change.}
The visualization results of background change are shown in Fig.~\ref{background change}, we can discover that Tune-A-Video, FateZero, and StableVideo fail to change the ``shopping and entertainment center'' into ``desolate martian landspace'' for the left sample.
In contrast, TCVE consistently changes the background according to the target prompt.
For the right sample, TCVE adeptly alters the background depicting the wave, whilst effectively preserving the original surfing postures of the individuals.

\textbf{Multiple-object editing.}
Besides the above single object editing, we also explore the challenging multiple-object editing task, as shown in Fig.~\ref{multiple}.
From these results, we observe that the all the methods except StableVideo can successfully change the background for the two illustrated samples. 
Nevertheless, Tune-A-Video and FateZero show evident shortcomings concerning the coherence of foreground objects. For the left sample, the astronaut and horse encounter a substantial reduction in visibility; For the right one, the car shows conspicuous inconsistency across consecutive frames.
In contrast, our proposed TCVE demonstrates the ability to produce videos with enhanced temporal coherence, showcasing proficient editing of both backgrounds and objects.

\subsection{Ablation Studies}

\subsubsection{Exploring effects of the key components in TCVE}
We evaluate the effects of the key components in TCVE, including Temporal Unet (TU) and the STU.
The results are illustrated in Tab.~\ref{tab:2}, please note that TCVE w/o STU means removing the STU and directly fusing $\X_{\text{spa}}$ and $\X_{\text{tem}}$ with a simple element-wise summation operation.
From these results, we can draw the following conclusions: \textbf{1)} TU is effective and brings considerable performance improvements in frame consistency due to its promising property for temporal modeling. \textbf{2)} STU brings further benefits, which proves that bridging T2I 2D Unet and TU using STU can further align the edited video with the target prompt while enhancing temporal consistency.

\begin{table}[!h]
	\centering
	\caption{Ablation studies of the key components in TCVE.}\label{tab:2}
	\setlength{\tabcolsep}{3.5pt}
	\scalebox{1.0}{
		\begin{tabular}{c|c|c|c}
			\hline
			Methods & Frame Consistency & Textual Alignment &  PickScore \\
			\hline
			\hline
			TCVE w/o TU & 90.65  & 27.34 & 20.42 \\
			TCVE w/o STU & 92.74  & 26.98 & 20.38 \\
			TCVE & \textbf{94.44} & \textbf{27.58} & \textbf{20.50} \\
			\hline
	\end{tabular}}
\end{table}

\begin{figure}[!h]
	\centering{\includegraphics[width=0.95\linewidth]{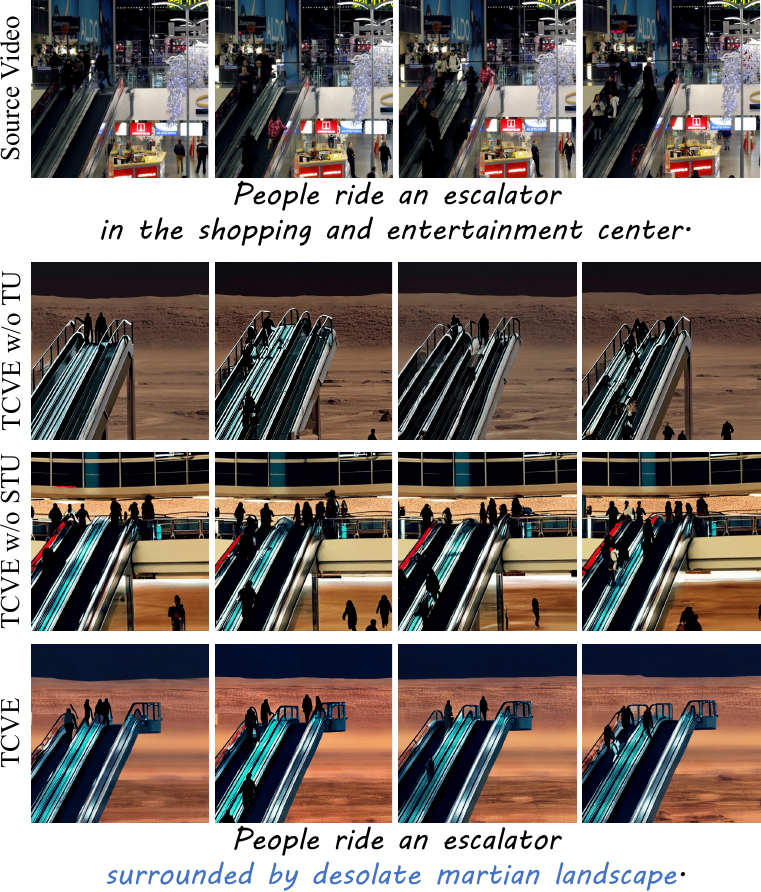}}
	\caption{Visualization results of ablation studies. TCVE w/o TU causes some flickering artifacts and inconsistency due to the fact that the escalator is deformed to varying degrees in different frames. TCVE w/o STU can generate smooth video frames, but it fails to achieve the purpose of the targeted editing. In contrast, TCVE could simultaneously maintain inter-frame consistency and editing capabilities.
	}
	\label{ablation}
\end{figure}

We provide the visualization results in Fig.~\ref{ablation} to further analyze the effects of each component. 
For TCVE w/o TU, there exists some flickering artifacts and inconsistency due to the fact that the escalator is deformed to varying degrees in different frames.
In contrast, TCVE preserves consistent structure for the escalator among frames, which proves the effectiveness of TU in modeling temporal information.
For TCVE w/o STU, the generated video frames are smooth. However, it fails to achieve the purpose of the targeted editing. This phenomenon suggests that directly fusing temporal-aware and spatial-aware features may lead to a compromise in the editing capabilities.
In contrast, TCVE demonstrates the ability to produce a seamlessly flowing video while effectively conveying the intended editing objectives. This substantiates the efficacy of STU in preserving both temporal coherence and the capacity for video editing.

\subsubsection{Exploring effects of the key components in STU}
Although the effectiveness of STU is verified in the previous part, the effectiveness of key components in STU remains to be explored.
Therefore, we now evaluate the effects of the key components in STU, including temporal attention (TA) and 3D convolution (3DConv).
The results are reported in Tab.~\ref{tab:3}, we observe the following conclusions: \textbf{1)} STU w/o TA degrades more frame consistency due to the lack of temporal awareness. \textbf{2)} STU w/o 3DConv reduces more textual alignment due to the lack of spatio-temporal modeling.
The above results demonstrate the effectiveness of TA and 3DConv in maintaining frame consistency and textual alignment, respectively.

\begin{table}[!h]
	\centering
	\caption{Ablation studies of the key components in STU.}\label{tab:3}
	\setlength{\tabcolsep}{3.5pt}
	\scalebox{1.0}{
		\begin{tabular}{c|c|c|c}
			\hline
			Methods & Frame Consistency & Textual Alignment &  PickScore \\
			\hline
			\hline
			STU w/o TA & 93.51  & 27.41 & 20.48 \\
			STU w/o 3DConv & 93.92  & 27.04 & 20.41 \\
			TCVE & \textbf{94.44} & \textbf{27.58} & \textbf{20.50} \\
			\hline
	\end{tabular}}
\end{table}

\section{Conclusion}
In this paper, we challenge the temporal inconsistency issue in text-guided video editing by proposing a straightforward and effective Temporal-Consistent Video Editing method.
To model the temporal information, we construct a temporal Unet model inspired by the pretrained T2I 2D Unet to facilitate temporal-focused modeling.
To bridge the temporal Unet and pretrained T2I 2D Unet, we design a spatial-temporal modeling unit to perceive both temporal and spatial information, thereby maintaining both the temporal consistency of video and the desired editing capability.
Quantitative and qualitative experiments prove the validity of TCVE.
A limitation is that our TCVE may cause some unsatisfactory results when simultaneously manipulating style, objects, and backgrounds.
This could be attributed to the fact that the text conditioning embedding stems from the CLIP text encoder, which aligns predominantly with image-based embeddings and may not seamlessly correspond with video samples.
A potential solution is to use an additional video-based CLIP model as text embedding.
This avenue of research is left as future work.

\bibliographystyle{IEEEtran}
\bibliography{refs}

% Generated by IEEEtran.bst, version: 1.14 (2015/08/26)
\begin{thebibliography}{10}
\providecommand{\url}[1]{#1}
\csname url@samestyle\endcsname
\providecommand{\newblock}{\relax}
\providecommand{\bibinfo}[2]{#2}
\providecommand{\BIBentrySTDinterwordspacing}{\spaceskip=0pt\relax}
\providecommand{\BIBentryALTinterwordstretchfactor}{4}
\providecommand{\BIBentryALTinterwordspacing}{\spaceskip=\fontdimen2\font plus
\BIBentryALTinterwordstretchfactor\fontdimen3\font minus
  \fontdimen4\font\relax}
\providecommand{\BIBforeignlanguage}[2]{{%
\expandafter\ifx\csname l@#1\endcsname\relax
\typeout{** WARNING: IEEEtran.bst: No hyphenation pattern has been}%
\typeout{** loaded for the language `#1'. Using the pattern for}%
\typeout{** the default language instead.}%
\else
\language=\csname l@#1\endcsname
\fi
#2}}
\providecommand{\BIBdecl}{\relax}
\BIBdecl

\bibitem{ddpm}
J.~Ho, A.~Jain, and P.~Abbeel, ``Denoising diffusion probabilistic models,''
  \emph{Advances in neural information processing systems}, vol.~33, pp.
  6840--6851, 2020.

\bibitem{scoresde}
Y.~Song, J.~Sohl-Dickstein, D.~P. Kingma, A.~Kumar, S.~Ermon, and B.~Poole,
  ``Score-based generative modeling through stochastic differential
  equations,'' in \emph{International Conference on Learning Representations},
  2021.

\bibitem{imder}
Y.~Wang, Y.~Li, and Z.~Cui, ``Incomplete multimodality-diffused emotion
  recognition,'' in \emph{Thirty-seventh Conference on Neural Information
  Processing Systems}, 2023.

\bibitem{DiffFashion}
S.~Cao, W.~Chai, S.~Hao, Y.~Zhang, H.~Chen, and G.~Wang, ``Difffashion:
  Reference-based fashion design with structure-aware transfer by diffusion
  models,'' \emph{IEEE Transactions on Multimedia}, pp. 1--13, 2023.

\bibitem{stablediffusion}
R.~Rombach, A.~Blattmann, D.~Lorenz, P.~Esser, and B.~Ommer, ``High-resolution
  image synthesis with latent diffusion models,'' in \emph{Proceedings of the
  IEEE/CVF conference on computer vision and pattern recognition}, 2022, pp.
  10\,684--10\,695.

\bibitem{imagen}
C.~Saharia, W.~Chan, S.~Saxena, L.~Li, J.~Whang, E.~L. Denton, K.~Ghasemipour,
  R.~Gontijo~Lopes, B.~Karagol~Ayan, T.~Salimans \emph{et~al.},
  ``Photorealistic text-to-image diffusion models with deep language
  understanding,'' \emph{Advances in Neural Information Processing Systems},
  vol.~35, pp. 36\,479--36\,494, 2022.

\bibitem{controlnet}
L.~Zhang, A.~Rao, and M.~Agrawala, ``Adding conditional control to
  text-to-image diffusion models,'' in \emph{Proceedings of the IEEE/CVF
  International Conference on Computer Vision}, 2023, pp. 3836--3847.

\bibitem{Dreambooth}
N.~Ruiz, Y.~Li, V.~Jampani, Y.~Pritch, M.~Rubinstein, and K.~Aberman,
  ``Dreambooth: Fine tuning text-to-image diffusion models for subject-driven
  generation,'' in \emph{Proceedings of the IEEE/CVF Conference on Computer
  Vision and Pattern Recognition}, 2023, pp. 22\,500--22\,510.

\bibitem{Imagenvideo}
J.~Ho, W.~Chan, C.~Saharia, J.~Whang, R.~Gao, A.~Gritsenko, D.~P. Kingma,
  B.~Poole, M.~Norouzi, D.~J. Fleet \emph{et~al.}, ``Imagen video: High
  definition video generation with diffusion models,'' \emph{arXiv preprint
  arXiv:2210.02303}, 2022.

\bibitem{Make-A-Video}
U.~Singer, A.~Polyak, T.~Hayes, X.~Yin, J.~An, S.~Zhang, Q.~Hu, H.~Yang,
  O.~Ashual, O.~Gafni \emph{et~al.}, ``Make-a-video: Text-to-video generation
  without text-video data,'' in \emph{The Eleventh International Conference on
  Learning Representations}, 2023.

\bibitem{Alignyourlatents}
A.~Blattmann, R.~Rombach, H.~Ling, T.~Dockhorn, S.~W. Kim, S.~Fidler, and
  K.~Kreis, ``Align your latents: High-resolution video synthesis with latent
  diffusion models,'' in \emph{Proceedings of the IEEE/CVF Conference on
  Computer Vision and Pattern Recognition}, 2023, pp. 22\,563--22\,575.

\bibitem{Animatediff}
Y.~Guo, C.~Yang, A.~Rao, Y.~Wang, Y.~Qiao, D.~Lin, and B.~Dai, ``Animatediff:
  Animate your personalized text-to-image diffusion models without specific
  tuning,'' \emph{arXiv preprint arXiv:2307.04725}, 2023.

\bibitem{TAV}
J.~Z. Wu, Y.~Ge, X.~Wang, W.~Lei, Y.~Gu, W.~Hsu, Y.~Shan, X.~Qie, and M.~Z.
  Shou, ``Tune-a-video: One-shot tuning of image diffusion models for
  text-to-video generation,'' in \emph{Proceedings of the IEEE/CVF
  International Conference on Computer Vision}, 2023.

\bibitem{Blendeddiffusion}
O.~Avrahami, D.~Lischinski, and O.~Fried, ``Blended diffusion for text-driven
  editing of natural images,'' in \emph{Proceedings of the IEEE/CVF Conference
  on Computer Vision and Pattern Recognition}, 2022, pp. 18\,208--18\,218.

\bibitem{Plug-and-play}
N.~Tumanyan, M.~Geyer, S.~Bagon, and T.~Dekel, ``Plug-and-play diffusion
  features for text-driven image-to-image translation,'' in \emph{Proceedings
  of the IEEE/CVF Conference on Computer Vision and Pattern Recognition}, 2023,
  pp. 1921--1930.

\bibitem{DDIM}
J.~Song, C.~Meng, and S.~Ermon, ``Denoising diffusion implicit models,'' in
  \emph{International Conference on Learning Representations}, 2021.

\bibitem{fatezero}
C.~Qi, X.~Cun, Y.~Zhang, C.~Lei, X.~Wang, Y.~Shan, and Q.~Chen, ``Fatezero:
  Fusing attentions for zero-shot text-based video editing,'' in
  \emph{Proceedings of the IEEE/CVF International Conference on Computer
  Vision}, 2023.

\bibitem{AlignDRAW}
E.~Mansimov, E.~Parisotto, J.~L. Ba, and R.~Salakhutdinov, ``Generating images
  from captions with attention,'' in \emph{International Conference on Learning
  Representations}, 2016.

\bibitem{t2i-icml}
S.~Reed, Z.~Akata, X.~Yan, L.~Logeswaran, B.~Schiele, and H.~Lee, ``Generative
  adversarial text to image synthesis,'' in \emph{International conference on
  machine learning}.\hskip 1em plus 0.5em minus 0.4em\relax PMLR, 2016, pp.
  1060--1069.

\bibitem{ControllableT2I}
B.~Li, X.~Qi, T.~Lukasiewicz, and P.~Torr, ``Controllable text-to-image
  generation,'' \emph{Advances in Neural Information Processing Systems},
  vol.~32, 2019.

\bibitem{DALL-E}
A.~Ramesh, M.~Pavlov, G.~Goh, S.~Gray, C.~Voss, A.~Radford, M.~Chen, and
  I.~Sutskever, ``Zero-shot text-to-image generation,'' in \emph{International
  Conference on Machine Learning}.\hskip 1em plus 0.5em minus 0.4em\relax PMLR,
  2021, pp. 8821--8831.

\bibitem{t2i-tmm1}
R.~Li, N.~Wang, F.~Feng, G.~Zhang, and X.~Wang, ``Exploring global and local
  linguistic representations for text-to-image synthesis,'' \emph{IEEE
  Transactions on Multimedia}, vol.~22, no.~12, pp. 3075--3087, 2020.

\bibitem{t2i-tmm2}
Q.~Cheng, K.~Wen, and X.~Gu, ``Vision-language matching for text-to-image
  synthesis via generative adversarial networks,'' \emph{IEEE Transactions on
  Multimedia}, 2022.

\bibitem{GANs}
I.~Goodfellow, J.~Pouget-Abadie, M.~Mirza, B.~Xu, D.~Warde-Farley, S.~Ozair,
  A.~Courville, and Y.~Bengio, ``Generative adversarial nets,'' \emph{Advances
  in neural information processing systems}, vol.~27, 2014.

\bibitem{gansurvvey}
A.~Creswell, T.~White, V.~Dumoulin, K.~Arulkumaran, B.~Sengupta, and A.~A.
  Bharath, ``Generative adversarial networks: An overview,'' \emph{IEEE signal
  processing magazine}, vol.~35, no.~1, pp. 53--65, 2018.

\bibitem{Glow}
D.~P. Kingma and P.~Dhariwal, ``Glow: Generative flow with invertible 1x1
  convolutions,'' \emph{Advances in neural information processing systems},
  vol.~31, 2018.

\bibitem{dicmor}
Y.~Wang, Z.~Cui, and Y.~Li, ``Distribution-consistent modal recovering for
  incomplete multimodal learning,'' in \emph{Proceedings of the IEEE/CVF
  International Conference on Computer Vision}, 2023, pp. 22\,025--22\,034.

\bibitem{Parti}
J.~Yu, Y.~Xu, J.~Y. Koh, T.~Luong, G.~Baid, Z.~Wang, V.~Vasudevan, A.~Ku,
  Y.~Yang, B.~K. Ayan \emph{et~al.}, ``Scaling autoregressive models for
  content-rich text-to-image generation,'' \emph{arXiv preprint
  arXiv:2206.10789}, vol.~2, no.~3, p.~5, 2022.

\bibitem{DALLE-2}
A.~Ramesh, P.~Dhariwal, A.~Nichol, C.~Chu, and M.~Chen, ``Hierarchical
  text-conditional image generation with clip latents,'' \emph{arXiv preprint
  arXiv:2204.06125}, 2022.

\bibitem{GLIDE}
A.~Q. Nichol, P.~Dhariwal, A.~Ramesh, P.~Shyam, P.~Mishkin, B.~Mcgrew,
  I.~Sutskever, and M.~Chen, ``Glide: Towards photorealistic image generation
  and editing with text-guided diffusion models,'' in \emph{International
  Conference on Machine Learning}.\hskip 1em plus 0.5em minus 0.4em\relax PMLR,
  2022, pp. 16\,784--16\,804.

\bibitem{Reco}
Z.~Yang, J.~Wang, Z.~Gan, L.~Li, K.~Lin, C.~Wu, N.~Duan, Z.~Liu, C.~Liu,
  M.~Zeng \emph{et~al.}, ``Reco: Region-controlled text-to-image generation,''
  in \emph{Proceedings of the IEEE/CVF Conference on Computer Vision and
  Pattern Recognition}, 2023, pp. 14\,246--14\,255.

\bibitem{clip}
A.~Radford, J.~W. Kim, C.~Hallacy, A.~Ramesh, G.~Goh, S.~Agarwal, G.~Sastry,
  A.~Askell, P.~Mishkin, J.~Clark \emph{et~al.}, ``Learning transferable visual
  models from natural language supervision,'' in \emph{International conference
  on machine learning}.\hskip 1em plus 0.5em minus 0.4em\relax PMLR, 2021, pp.
  8748--8763.

\bibitem{Spatext}
O.~Avrahami, T.~Hayes, O.~Gafni, S.~Gupta, Y.~Taigman, D.~Parikh,
  D.~Lischinski, O.~Fried, and X.~Yin, ``Spatext: Spatio-textual representation
  for controllable image generation,'' in \emph{Proceedings of the IEEE/CVF
  Conference on Computer Vision and Pattern Recognition}, 2023, pp.
  18\,370--18\,380.

\bibitem{VDM}
J.~Ho, T.~Salimans, A.~Gritsenko, W.~Chan, M.~Norouzi, and D.~J. Fleet, ``Video
  diffusion models,'' \emph{Advances in Neural Information Processing Systems},
  2022.

\bibitem{Magicvideo}
D.~Zhou, W.~Wang, H.~Yan, W.~Lv, Y.~Zhu, and J.~Feng, ``Magicvideo: Efficient
  video generation with latent diffusion models,'' \emph{arXiv preprint
  arXiv:2211.11018}, 2022.

\bibitem{higen}
Z.~Qing, S.~Zhang, J.~Wang, X.~Wang, Y.~Wei, Y.~Zhang, C.~Gao, and N.~Sang,
  ``Hierarchical spatio-temporal decoupling for text-to-video generation,''
  \emph{arXiv preprint arXiv:2312.04483}, 2023.

\bibitem{Modelscope}
J.~Wang, H.~Yuan, D.~Chen, Y.~Zhang, X.~Wang, and S.~Zhang, ``Modelscope
  text-to-video technical report,'' \emph{arXiv preprint arXiv:2308.06571},
  2023.

\bibitem{Pvdm}
S.~Ge, S.~Nah, G.~Liu, T.~Poon, A.~Tao, B.~Catanzaro, D.~Jacobs, J.-B. Huang,
  M.-Y. Liu, and Y.~Balaji, ``Preserve your own correlation: A noise prior for
  video diffusion models,'' in \emph{Proceedings of the IEEE/CVF International
  Conference on Computer Vision}, 2023, pp. 22\,930--22\,941.

\bibitem{Text2video-zero}
L.~Khachatryan, A.~Movsisyan, V.~Tadevosyan, R.~Henschel, Z.~Wang,
  S.~Navasardyan, and H.~Shi, ``Text2video-zero: Text-to-image diffusion models
  are zero-shot video generators,'' \emph{arXiv preprint arXiv:2303.13439},
  2023.

\bibitem{DSNET}
B.~Liu, X.~Liu, A.~Dai, Z.~Zeng, Z.~Cui, and J.~Yang, ``Dual-stream diffusion
  net for text-to-video generation,'' \emph{arXiv preprint arXiv:2308.08316},
  2023.

\bibitem{SDEdit}
C.~Meng, Y.~He, Y.~Song, J.~Song, J.~Wu, J.-Y. Zhu, and S.~Ermon, ``Sdedit:
  Guided image synthesis and editing with stochastic differential equations,''
  in \emph{International Conference on Learning Representations}, 2021.

\bibitem{DiffEdit}
G.~Couairon, J.~Verbeek, H.~Schwenk, and M.~Cord, ``Diffedit: Diffusion-based
  semantic image editing with mask guidance,'' in \emph{The Eleventh
  International Conference on Learning Representations}, 2023.

\bibitem{Imagic}
B.~Kawar, S.~Zada, O.~Lang, O.~Tov, H.~Chang, T.~Dekel, I.~Mosseri, and
  M.~Irani, ``Imagic: Text-based real image editing with diffusion models,'' in
  \emph{Proceedings of the IEEE/CVF Conference on Computer Vision and Pattern
  Recognition}, 2023, pp. 6007--6017.

\bibitem{text2live}
O.~Bar-Tal, D.~Ofri-Amar, R.~Fridman, Y.~Kasten, and T.~Dekel, ``Text2live:
  Text-driven layered image and video editing,'' in \emph{European Conference
  on Computer Vision}.\hskip 1em plus 0.5em minus 0.4em\relax Springer, 2022,
  pp. 707--723.

\bibitem{gen1}
P.~Esser, J.~Chiu, P.~Atighehchian, J.~Granskog, and A.~Germanidis, ``Structure
  and content-guided video synthesis with diffusion models,'' \emph{arXiv
  preprint arXiv:2302.03011}, 2023.

\bibitem{loeschcke2022text}
S.~Loeschcke, S.~Belongie, and S.~Benaim, ``Text-driven stylization of video
  objects,'' in \emph{European Conference on Computer Vision}.\hskip 1em plus
  0.5em minus 0.4em\relax Springer, 2022, pp. 594--609.

\bibitem{stablevideo}
W.~Chai, X.~Guo, G.~Wang, and Y.~Lu, ``Stablevideo: Text-driven
  consistency-aware diffusion video editing,'' in \emph{Proceedings of the
  IEEE/CVF International Conference on Computer Vision}, 2023, pp.
  23\,040--23\,050.

\bibitem{atlases}
Y.~Kasten, D.~Ofri, O.~Wang, and T.~Dekel, ``Layered neural atlases for
  consistent video editing,'' \emph{ACM Transactions on Graphics (TOG)},
  vol.~40, no.~6, pp. 1--12, 2021.

\bibitem{dreamix}
E.~Molad, E.~Horwitz, D.~Valevski, A.~R. Acha, Y.~Matias, Y.~Pritch,
  Y.~Leviathan, and Y.~Hoshen, ``Dreamix: Video diffusion models are general
  video editors,'' \emph{arXiv preprint arXiv:2302.01329}, 2023.

\bibitem{attention}
A.~Vaswani, N.~Shazeer, N.~Parmar, J.~Uszkoreit, L.~Jones, A.~N. Gomez,
  L.~Kaiser, and I.~Polosukhin, ``Attention is all you need,'' \emph{Advances
  in neural information processing systems}, vol.~30, 2017.

\bibitem{p2p}
A.~Hertz, R.~Mokady, J.~Tenenbaum, K.~Aberman, Y.~Pritch, and D.~Cohen-Or,
  ``Prompt-to-prompt image editing with cross attention control,'' \emph{arXiv
  preprint arXiv:2208.01626}, 2022.

\bibitem{Nulltext}
R.~Mokady, A.~Hertz, K.~Aberman, Y.~Pritch, and D.~Cohen-Or, ``Null-text
  inversion for editing real images using guided diffusion models,'' in
  \emph{Proceedings of the IEEE/CVF Conference on Computer Vision and Pattern
  Recognition}, 2023, pp. 6038--6047.

\bibitem{unet}
O.~Ronneberger, P.~Fischer, and T.~Brox, ``U-net: Convolutional networks for
  biomedical image segmentation,'' in \emph{Medical Image Computing and
  Computer-Assisted Intervention--MICCAI 2015: 18th International Conference,
  Munich, Germany, October 5-9, 2015, Proceedings, Part III 18}.\hskip 1em plus
  0.5em minus 0.4em\relax Springer, 2015, pp. 234--241.

\bibitem{pickscore}
Y.~Kirstain, A.~Polyak, U.~Singer, S.~Matiana, J.~Penna, and O.~Levy,
  ``Pick-a-pic: An open dataset of user preferences for text-to-image
  generation,'' \emph{arXiv preprint arXiv:2305.01569}, 2023.

\end{thebibliography}

\vfill

\end{document}